\pdfoutput=1

\documentclass[11pt]{article}

\usepackage{emnlp2021}

\usepackage{times}
\usepackage{latexsym}

\usepackage[T1]{fontenc}

\usepackage[utf8]{inputenc}
\usepackage{xcolor,colortbl}

\usepackage{microtype}
\usepackage{times}
\usepackage{latexsym}
\usepackage{float}
\usepackage{graphicx}
\usepackage{booktabs}
\usepackage{amsmath,amsfonts,amssymb,amsthm}
\usepackage{bm}
\usepackage[ruled,noend]{algorithm2e}
\usepackage{enumitem}
\usepackage{setspace}
\usepackage{multirow}

\usepackage{cleveref}
\crefformat{section}{Sec.~#2#1#3}
\crefformat{subsection}{Sec.~#2#1#3}
\crefformat{subsubsection}{Sec.~#2#1#3}
\crefrangeformat{section}{Sec.#3#1#4 to~#5#2#6}
\crefrangeformat{subsection}{Sec.#3#1#4 to~#5#2#6}
\crefmultiformat{section}{Sec.#2#1#3}{ and~#2#1#3}{, #2#1#3}{ and~#2#1#3}
\usepackage{refstyle}
\Crefformat{figure}{#2Fig.~#1#3}
\Crefmultiformat{figure}{Figs.~#2#1#3}{ and~#2#1#3}{, #2#1#3}{ and~#2#1#3}
\Crefformat{table}{#2Tab.~#1#3}
\Crefmultiformat{table}{Tabs.~#2#1#3}{ and~#2#1#3}{, #2#1#3}{ and~#2#1#3}
\Crefformat{appendix}{Appx.~#2#1#3}
\crefformat{algorithm}{Alg.~#2#1#3}

\Crefformat{equation}{Eq.~#2#1#3}

\newcommand{\stitle}[1]{\vspace{0.3em}\noindent{\bf #1}}
\newcommand{\colorit}{\cellcolor{green!15}}
\SetKwInput{KwInput}{Input}
\SetKwInput{KwOutput}{Output}

\definecolor{plot_orange}{rgb}{1.0, 0.498, 0.055}
\definecolor{plot_blue}{rgb}{0.121, 0.467, 0.706}
\definecolor{plot_grey}{rgb}{0.501, 0.501, 0.501}

\title{Contrastive Out-of-Distribution Detection for Pretrained Transformers}

\author{Wenxuan Zhou \\
  University of Southern California \\
  \texttt{zhouwenx@usc.edu} \\\And
  Fangyu Liu \\
  University of Cambridge \\
  \texttt{fl399@cam.ac.uk} \\\And
  Muhao Chen \\
  University of Southern California \\
  \texttt{muhaoche@usc.edu} \\}

\begin{document}
\maketitle

\begin{abstract}
Pretrained Transformers achieve remarkable performance when training and test data are 
from the same distribution.
However, in real-world scenarios, the model often faces out-of-distribution~(OOD) instances 
that can cause severe semantic shift problems at inference time.
Therefore, in practice, a reliable model should identify such instances, and then either reject them during inference or pass them over to 
models that handle another distribution.
In this paper, we develop an unsupervised OOD detection method, in which only the in-distribution~(ID) data are used in training.
We propose to fine-tune the Transformers with a contrastive loss, which improves the compactness of representations, such that OOD instances can be better differentiated from ID ones.
These OOD instances can then be accurately detected using the Mahalanobis distance in the model's penultimate layer.
We experiment with comprehensive settings and achieve near-perfect OOD detection performance, outperforming baselines drastically. We further investigate the rationales behind the improvement, finding that more compact representations through margin-based contrastive learning bring the improvement. 
We release our code to the community for future research\footnote{The implementation and datasets are available at \url{https://github.com/wzhouad/Contra-OOD}}.
\end{abstract}

\section{Introduction}
Many natural language 
classifiers are developed based on a closed-world assumption, i.e., the training and test data are sampled from the same distribution.
However, training data can rarely capture the entire distribution.
In real-world scenarios, out-of-distribution~(OOD) instances, which come from categories that are not known to the model, can often be present in inference phases.
These instances could be misclassified by the model into known categories with high confidence, 
causing the semantic shift problem 
\cite{Hsu2020GeneralizedOD}.
As a practical solution to this problem in real-world applications, the model should detect such instances, and signal exceptions or transmit to models handling other categories or tasks.
Although pretrained Transformers~\cite{devlin-etal-2019-bert} achieve remarkable results when intrinsically evaluated on in-distribution (ID) data, recent work~\cite{hendrycks-etal-2020-pretrained} shows that many of these models fall short of detecting OOD instances.

Despite the importance, few attempts have been made for the problem of detecting OOD in NLP tasks.
One proposed method is to train a model on both the ID and OOD data and regularize the model to produce lower confidence on OOD instances than ID ones~\cite{Hendrycks2019DeepAD,Larson2019AnED}.
However, as the OOD instances reside in an unbounded feature space, their distribution during inference is usually unknown.
Hence, it is hard to decide which OOD instances to use in training, let alone that they may not be available in lots of scenarios.
Another practiced method for OOD detection is to use the maximum class probability as an indicator~\cite{Shu2017DOCDO,hendrycks-etal-2020-pretrained}, 
such that lower values indicate more probable OOD instances.
Though easy to implement, its OOD detection performance is far from perfection, as
prior studies~\cite{Dhamija2018ReducingNA,liang2018enhancing} show that OOD inputs can often get high probabilities as well. 

In this paper, we aim at improving the OOD detection ability of natural language classifiers, in particular, the pretrained Transformers, which have been the backbones of many SOTA NLP systems.
For practical purposes, we adopt the setting where only ID data are available during task-specific training.
Moreover, we require that the model should maintain classification performance on the ID task data.
To this end, we propose a contrastive learning framework for unsupervised OOD detection, which is composed of a contrastive loss and an OOD scoring function.
Our contrastive loss aims at increasing the discrepancy of the representations of instances from different classes in the task.
During training, instances belonging to the same class are regarded as pseudo-ID data while those of different classes are considered mutually pseudo-OOD data.
We hypothesize that increasing inter-class discrepancies can help the model learn discriminative features for ID/OOD distinctions, and therefore help detect true OOD data at inference.
We study two versions of the contrastive loss: a similarity-based contrastive loss~\cite{sohn2016improved,oord2018representation,chen2020simple} and a margin-based contrastive loss.
The OOD scoring function maps the representations of instances to OOD detection scores, indicating the likelihood of an instance being OOD.
We examine different combinations of contrastive losses and OOD scoring functions, including maximum softmax probability, energy score, Mahalanobis distance, and maximum cosine similarity.
Particularly, we observe that OOD scoring based on the Mahalanobis distance~\cite{lee2018simple}, when incorporated with the margin-based contrastive loss, generally leads to the best OOD detection performance.
The Mahalanobis distance is computed from the penultimate layer\footnote{I.e., the input to the softmax layer.} of Transformers by fitting a class-conditional multivariate Gaussian distribution.

The main contributions of this work are three-fold.
First, we propose a contrastive learning framework for unsupervised OOD detection, where we comprehensively study combinations of different contrastive learning losses and OOD scoring functions.
Second, extensive experiments on various tasks and datasets demonstrate the significant improvement our method has made to OOD detection for Transformers.
Third, we provide a detailed analysis to reveal the importance of different incorporated techniques, which also identifies further challenges for this emerging research topic.

\section{Related Work}

\stitle{Out-of-Distribution Detection.}
Determining whether an instance is OOD is critical for the safe deployment of machine learning systems in the real world~\cite{Amodei2016ConcretePI}.
The main challenge 
is that the distribution of OOD data is hard to estimate \textit{a priori}.
Based on the availability of OOD data, recent methods can be categorized into supervised, self-supervised, and unsupervised ones.
Supervised methods train models on both ID and OOD data, where the models are expected to output a uniform distribution over known classes on OOD data~\cite{Lee2018TrainingCC,Dhamija2018ReducingNA,Hendrycks2019DeepAD}.
However, it is hard to assume the presence of a large dataset that provides comprehensive coverage for OOD instances in practice.
Self-supervised methods~\cite{bergman2019classification} apply augmentation techniques to change certain properties of data (e.g., through rotation of an image) and simultaneously learn an auxiliary model to predict the property changes (e.g., the rotation angle).
Such an auxiliary model is expected to have worse generalization on OOD data which can in turn be identified by a larger loss.
However, it is hard to define such transformations for natural language.
Unsupervised methods use only ID data in training.
They detect OOD data based on the class probabilities~\cite{Bendale2016TowardsOS,hendrycks2016baseline,Shu2017DOCDO,liang2018enhancing} 
or other latent space metrics~\cite{Liu2020EnergybasedOD,lee2018simple}.
Particularly, \citet{vyas2018out} randomly split the training classes into two subsets and treat them as pseudo-ID and pseudo-OOD data, respectively.
They then train an OOD detector that requires the entropy of probability distribution on pseudo-OOD data to be lower than pseudo-ID data.
This process is repeated to obtain multiple OOD detectors, and their ensemble is used to detect the OOD instances.
This method conducts OOD detection at the cost of high computational overhead in training redundant models and has the limitation of not supporting the detection for binary classification tasks.


Though extensively studied for computer vision (CV), OOD detection has been overlooked in NLP,
and most prior works~\cite{kim2018joint,Hendrycks2019DeepAD,tan-etal-2019-domain} require both ID and OOD data in training. 
\citet{hendrycks-etal-2020-pretrained} use the maximum softmax probability as the detection score and show that pretrained Transformers exhibit better OOD detection performance than models such as LSTM~\cite{Hochreiter1997LongSM}, while the performance is still imperfect.
Our framework, as an unsupervised OOD detection approach, significantly improves the OOD detection of Transformers only using ID data.

\stitle{Contrastive Learning.}
Recently, contrastive learning has received a lot of research attention.
It works by mapping instances of the same class into a nearby region and make instances of different classes uniformly distributed~\cite{wang2020understanding}. 
Many efforts on CV~\cite{misra2020self,he2020momentum,chen2020simple} and NLP~\cite{giorgi2020declutr} incorporate contrastive learning into self-supervised learning, which seeks to gather the representations of different augmented views of the same instance and separate those of different instances.
Prior work on image classification~\cite{Tack2020CSIND,winkens2020contrastive} shows that model trained with self-supervised contrastive learning generates discriminative features for detecting distributional shifts.
However, such methods heavily rely on data augmentation of instances and are hard to be applied to NLP.
Other efforts on CV~\cite{khosla2020supervised} and NLP~\cite{gunel2020supervised} conduct contrastive learning in a supervised manner, which aims at 
embedding instances of the same class closer and separating different classes.
They show that models trained with supervised contrastive learning exhibit better classification performance.
To the best of our knowledge, we are the first to introduce supervised contrastive learning to OOD detection.
Such a method does not rely on data augmentation, thus can be easily adapted to existing NLP models. We also propose a margin-based contrastive objective that greatly outperforms standard supervised contrastive losses.

\section{Method}
In this section, we first formally define the OOD detection problem (\Cref{ssec:task_definition}), then introduce the overall framework (\Cref{secc:overview}), and finally present the contrastive representation learning and scoring functions (\Cref{sec:contrastive} and \Cref{sec:scorer}).

\subsection{Problem Definition}\label{ssec:task_definition}
We aim at improving the OOD detection performance of natural language classifiers that are based on pretrained Transformers, 
using only ID data in the main-task training.
Generally, the out-of-distribution~(OOD) instances can be defined as instances $(\bm{x}, y)$ sampled from an underlying distribution other than the training distribution $\mathrm{P}(\mathcal{X}_\text{train}, \mathcal{Y}_\text{train})$, where $\mathcal{X}_\text{train}$ and $\mathcal{Y}_\text{train}$ are the training corpus and training label set, respectively.
In this context, literature further divides OOD data into those with \emph{semantic shift} or \emph{non-semantic shift}~\cite{Hsu2020GeneralizedOD}.
\emph{Semantic shift} refers to the instances that do not belong to $\mathcal{Y}_\text{train}$.
More specifically, instances with semantic shift may come from unknown categories or irrelevant tasks.
Therefore, the model is expected to detect and reject such instances (or forward them to models handling other tasks), instead of mistakenly classifying them into $\mathcal{Y}_\text{train}$.
\emph{Non-semantic shift}, on the other hand, refers to the instances that belong to $\mathcal{Y}_\text{train}$ but are sampled from a distribution other than $\mathcal{X}_\text{train}$, e.g., a different corpus.
Though drawn from OOD, those instances can be classified into $\mathcal{Y}_\text{train}$, thus can be accepted by the model.
Hence, in the context of this paper, we primarily consider an instance $(\bm{x}, y)$ to be OOD if $y\notin \mathcal{Y}_\text{train}$, i.e., exhibiting semantic shift, to be consistent with the problem settings of prior studies~\cite{hendrycks2016baseline,lee2018simple,hendrycks-etal-2020-pretrained}.

We hereby formally define the OOD detection task.
Specifically, given a main task of natural language classification (e.g., sentence classification, NLI, etc.), for an instance $\bm{x}$ to be classified, our goal is to develop an auxiliary OOD scoring function $f(\bm{x}): \mathcal{X} \rightarrow \mathbb{R}$. 
This function should return a low score for an ID instance where $y \in \mathcal{Y}_\text{train}$, and a high score for an OOD instance where $y \notin \mathcal{Y}_\text{train}$ ($y$ is the underlying label for $\bm{x}$ and is unknown at inference).
During inference, we can set a threshold for the OOD score 
to filter out most OOD instances.
This process involves a trade-off between false negative and false positive and may be specific to the application.
Meanwhile, we expect that the OOD detection auxiliary should not negatively 
affect the performance of the main task on ID data.

\subsection{Framework Overview}\label{secc:overview}
Next, we introduce the 
formation of our contrastive learning framework for OOD detection.
We decompose OOD detection into two steps.
The first step is contrastive representation learning, where we focus on learning a representation space $\mathcal{H}$ where the distribution of ID and that of OOD data are distinct.
Accordingly, we need another function to map the representation to an OOD score.
This process is equivalent to expressing OOD detection as $f(\bm{x}) = g(\bm{h})$,
where $\bm{h} \in \mathcal{H}$ is the dense representation of the input text $\bm{x}$ given by an encoder, $g: \mathcal{H} \rightarrow \mathbb{R}$ is a scoring function mapping the representation to an OOD detection score.
Using this decomposition, we can use different training strategies for $\bm{h}$ and different functions for $g$, which are studies in the following sections.

{
\begin{algorithm}[!t]
    \caption{Learning Process}\label{algo::main}
    \small
    \KwInput{ID training set $\mathcal{D}_\text{train}$ and ID validation set $\mathcal{D}_\text{val}$.}
    \KwOutput{A trained classifier and an OOD detector.}
    Initialize the pretrained Transformer $M$. \\
    \For {$t=1...T$}{
    Sample a batch from $\mathcal{D}_\text{train}$. \\
    Calculate the classification loss $\mathcal{L}_\text{ce}$. \\
    Calculate the contrastive loss $\mathcal{L}_\text{cont}$ 
    as either $\mathcal{L}_\text{scl}$ or $\mathcal{L}_\text{margin}$ \\
    $\mathcal{L}$ = $\mathcal{L}_\text{ce}$ + $\lambda \mathcal{L}_\text{cont}$. \\
    Update model parameters w.r.t. $\mathcal{L}$. \\
    \If{$t \; \mathsmaller{\%} \; \text{evaluation steps} = 0$ }{
        Fit the OOD detector on $\mathcal{D}_\text{val}$. \\
        Evaluate both the classifier and OOD detector on $\mathcal{D}_\text{val}$. \\
    }
    }
    Return the best model checkpoint. \\
\end{algorithm}
}

The learning process of our framework is described in \Cref{algo::main}.
In the training phase, our framework takes training and validation datasets that are both ID as input.
The model is optimized with both the (main task) classification loss and the contrastive loss on batches sampled from ID training data.
The best model is selected based on the ID validation data.
Specifically, for a distribution-based OOD scoring function such as the Mahalanobis distance, we first need to fit the OOD detector on the ID validation data.
We then evaluate the trained model on the ID validation data, where a satisfactory model should have a low contrastive loss and preserve the classification performance.
In the end, our framework returns a classifier to handle the main task on ID data and an OOD detector to identify OOD instances at inference.

\subsection{Contrastive Representation Learning}\label{sec:contrastive}
In this section, we discuss how to learn distinctive representations for OOD detection.
For better OOD detection performance, the representation space $\mathcal{H}$ is supposed to minimize the overlap of the representations of ID and OOD data.
In a supervised setting where both ID and OOD data are available in training, it would be easy to obtain such $\mathcal{H}$.
For example, \citet{Dhamija2018ReducingNA} train the neural model on both ID and OOD data and require the magnitude of representations of OOD instances to be smaller than ID representations.
However, in real-world applications, the distribution of OOD data is usually unknown beforehand.
We thus tackle a more general problem setting where the OOD data are assumed unavailable in training (unsupervised OOD detection, introduced below). 
\begin{figure}[!t]
    \centering
    \scalebox{0.39}{
    \includegraphics[width=\textwidth]{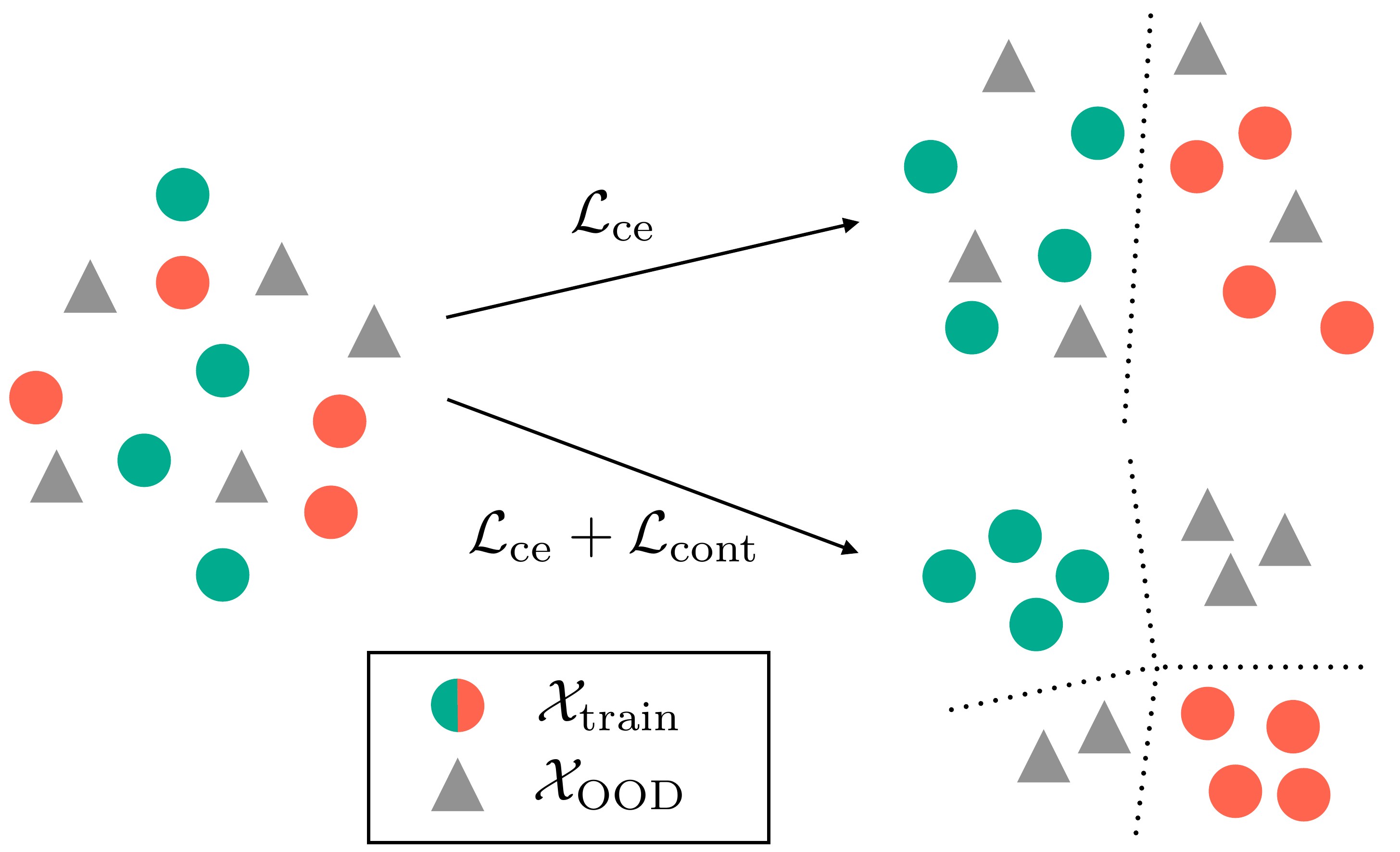}}
    \caption{Illustration of our proposed contrastive loss. The contrastive loss seeks to increase the discrepancy of the representations for instances from different training classes, such that OOD instances from unknown classes can be better differentiated.}
    \label{fig:illustration}
\end{figure}

In this unsupervised setup, though all training data used are ID, they may belong to different classes. We leverage data of distinct classes to learn more discriminative features. Through a contrastive learning objective, instances of the same class form compact clusters, while instances of different classes are encouraged to live apart from each other beyond a certain margin, as illustrated in \Cref{fig:illustration}. The discriminative feature space is generalizable to OOD data, which ultimately leads to better OOD detection performance in inference when encountering an unknown distribution.
We realize such a strategy using two alternatives of contrastive losses, i.e., the \emph{supervised contrastive loss} and the \emph{margin-based contrastive loss}.

\stitle{Supervised Contrastive Loss.}
Different from the contrastive loss used in self-supervised representation learning~\cite{chen2020simple,he2020momentum} that compares augmented instances to other instances, our contrastive loss contrasts instances to those from different ID classes.
To give a more specific illustration of our technique,
we first consider the supervised contrastive loss~\cite{khosla2020supervised,gunel2020supervised}.
Specifically, for a multi-class classification problem with $C$ classes, given a batch of training instances $\{(\bm{x}_i, y_i)\}_{i=1}^M$, where $\bm{x}_i$ is the input text, $y_i$ is the ground-truth label, the supervised contrastive loss can be formulated as:
\begin{equation*}\label{eq:scl}
    \mathcal{L}_\text{scl} = \sum_{i=1}^M \frac{-1}{M|P(i)|} \sum_{p\in P(i)} \log \frac{e^{\bm{z}_i^\intercal \bm{z}_p / \tau}}{\sum\limits_{a \in A(i)} e^{\bm{z}_i ^\intercal \bm{z}_a / \tau}},
\end{equation*}
where $A(i) = \{1,...,M\} \backslash \{i\}$ is the set of all anchor instances, $P(i) = \{p \in A(i): y_i = y_p\}$ is the set of anchor instances from the same class as $i$, $\tau$ is a temperature hyper-parameter, $\bm{z}$ is the L2-normalized \texttt{[CLS]} embedding before the softmax layer~\cite{khosla2020supervised,gunel2020supervised}.
The L2 normalization is for avoiding huge values in the dot product, which may lead to unstable updates.
In this case, this loss is optimized to increase the cosine similarity of instance pairs if they are from the same class and decrease it otherwise.

\stitle{Margin-based Contrastive Loss.}
The supervised contrastive loss produces minimal gradients when the similarity difference of positive and negative instances exceeds a certain point.
However, to better separate OOD instances, 
it is beneficial to enlarge the discrepancy between classes as much as possible.
Therefore, we propose another margin-based contrastive loss.
It encourages the L2 distances of instances from the same class to be as small as possible,  forming compact clusters, and the L2 distances of instances from different classes to be larger than a margin.
Our loss is formulated as:
\begin{align*}\label{eq:margin}
    &\mathcal{L}_\text{pos} = \sum_{i=1}^M \frac{1}{|P(i)|}\sum_{p\in P(i)} \Vert \bm{h}_i - \bm{h}_p \Vert^2, \nonumber \\
    &\mathcal{L}_\text{neg} = \sum_{i=1}^M \frac{1}{|N(i)|}\sum_{n\in N(i)} (\xi - \Vert \bm{h}_i - \bm{h}_n \Vert^2)_+, \nonumber \\
    &\mathcal{L}_\text{margin} = \frac{1}{dM} \left(\mathcal{L}_\text{pos} + \mathcal{L}_\text{neg}\right).
\end{align*}
Here $N(i)=\{n \in A(i): y_i \neq y_n\}$ is the set of anchor instances from other classes than $y_i$, $\bm{h} \in \mathbb{R}^d$ is the unnormalized \texttt{[CLS]} embedding before the softmax layer, $\xi$ is a margin, $d$ is the number of dimensions of $\bm{h}$.
As we do not use OOD data in training, it is hard to properly tune the margin.
Hence, we further incorporate an adaptive margin.
Intuitively, distances between instances from the same class should be smaller than those from different classes.
Therefore, we define the margin as the maximum distance between pairs of instances from the same class in the batch:
\begin{align*}
    \xi = \max_{i=1}^M \max_{p \in P(i)} \Vert \bm{h}_i - \bm{h}_p \Vert^2.
\end{align*}
We evaluate both contrastive losses in experiments.
In training, the model is jointly optimized with the cross-entropy classification loss $\mathcal{L}_\text{ce}$ and the contrastive loss $\mathcal{L}_\text{cont}$:
\begin{align*}
    \mathcal{L} = \mathcal{L}_\text{ce} + \lambda \mathcal{L}_\text{cont},
\end{align*}
where $\lambda$ is a positive coefficient.
We tune $\lambda$ based on the contrastive loss and the classification performance on the ID validation set, where a selected value for $\lambda$ should achieve a smaller contrastive loss while maintaining the classification performance.

\subsection{OOD Scoring Functions}\label{sec:scorer}
Next, we introduce 
the modeling of the OOD scoring function $g$.
The goal of the scoring function $g$ is to map the representations of instances to OOD detection scores, where higher scores indicate higher likelihoods for being OOD.
In the following, we describe several choices of this scoring function.

\stitle{Maximum Softmax Probability (MSP).} \citet{hendrycks2016baseline} use the maximum class probability $1 - \max_{j=1}^C \bm{p}_j$ among $C$ training classes in the softmax layer as an OOD indicator. 
This method has been widely adopted as a baseline for OOD detection~\cite{hendrycks2016baseline,Hsu2020GeneralizedOD,bergman2019classification,hendrycks-etal-2020-pretrained}.

\stitle{Energy Score (Energy).} \citet{Liu2020EnergybasedOD} interpret the softmax function as the ratio of the joint probability in $\mathcal{X} \times \mathcal{Y}$ to the probability in $\mathcal{X}$, and estimates the probability density of inputs as:
\begin{align*}
    g = -\log \sum_{j=1}^C \exp(\bm{w}_j^\intercal \bm{h}),
\end{align*}
where $\bm{w}_j \in \mathbb{R}^{d}$ is the weight of the $j^{th}$ class in the softmax layer, $\bm{h}$ is the input to the softmax layer.
A higher $g$ means lower probability density in ID classes and thus implies higher OOD likelihood.

\stitle{Mahalanobis Distance (Maha).} \citet{lee2018simple} model the ID features with class-conditional multivariate Gaussian distributions.
It first fits the Gaussian distributions on the ID validation set $\mathcal{D}_\text{val}=\{(\bm{x}_i, y_i)\}_{i=1}^M$ using the input representation $\bm{h}$ in the penultimate layer of model:
\begin{align*}
    \bm{\mu}_j &= \mathbb{E}_{y_i=j} \left[ \bm{h}_i \right], j=1, ..., C,\\
    \bm{\Sigma} &= \mathbb{E}\left[\left(\bm{h}_i - \bm{\mu}_{y_i}\right) \left(\bm{h}_i - \bm{\mu}_{y_i}\right)^\intercal\right],
\end{align*}
where $C$ is the number of classes, $\bm{\mu}_j$ is the mean vector of classes, and $\bm{\Sigma}$ is a shared covariance matrix of all classes.
Then, given an instance $\bm{x}$ during inference, it calculates the OOD detection score as the minimum Mahalanobis distance among the $C$ classes:
\begin{equation*}
    g = -\min_{j=1}^C (\bm{h} - \bm{\mu}_j)^\intercal \bm{\Sigma}^{+} (\bm{h} - \bm{\mu}_j),
\end{equation*}
where $\bm{\Sigma^+}$ is the pseudo-inverse of $\bm{\Sigma}$.
The Mahalanobis distance calculates the probability density of $\bm{h}$ in the Gaussian distribution.

\stitle{Cosine Similarity} can also be incorporated to consider the angular similarity of input representations.
To do so, 
the scoring function returns the OOD score as the maximum cosine similarity of $\bm{h}$ to instances of the ID validation set:
\begin{equation*}
    g = -\max_{i=1}^M\cos (\bm{h}, \bm{h}_i^{(val)}).
\end{equation*}

The above OOD scoring functions, combined with options of contrastive losses, lead to different variants of our framework.
We evaluate each combination in experiments.

\section{Experiments}

This section presents experimental evaluations of the proposed OOD detection framework. We start by describing experimental datasets and settings (\Cref{sec:dataset,sec:exp_set}), followed by detailed results analysis and case studies (\Cref{sec:results,secc:novel_class,sec:analysis}).

\subsection{Datasets}\label{sec:dataset}
Previous studies on OOD detection mostly focus on image classification, while few have been made on natural language.
Currently, there still lacks a well-established benchmark for OOD detection in NLP.
Therefore, we extend the selected datasets by~\citet{hendrycks-etal-2020-pretrained} and propose a more extensive benchmark, where we use different pairs of NLP datasets as ID and OOD data, respectively.
The criterion for dataset selection is that the OOD instances should not belong to ID classes.
To ensure this, we refer to the label descriptions in datasets and manually inspect samples of instances.

We use the following datasets as alternatives of ID data that correspond to three natural language classification tasks:

\begin{itemize}[leftmargin=1em]
    \setlength\itemsep{0em}
    \item \textbf{Sentiment Analysis.} We include two datasets for this task. \emph{SST2} \cite{Socher2013RecursiveDM} and \emph{IMDB} \cite{maas2011learning} are both datasets for sentiment analysis, where the polarities of sentences are labeled either positive or negative.
    For SST2, the train/validation/test splits are provided in the dataset.
    For IMDB, we randomly sample $10\%$ of the training instances as the validation set. Note that both datasets belong to the same task and are not considered OOD to each other.
    \item \textbf{Topic Classification.} We use \emph{20 Newsgroup} \cite{lang1995newsweeder}, a dataset for topic classification containing 20 classes.
    We randomly divide the whole dataset into an 80/10/10 split as the train/validation/test set.
    \item \textbf{Question Classification.} \emph{TREC-10} \cite{li2002learning} classifies questions based on the types of their sought-after answers.
    We use its coarse version with 6 classes and randomly sample $10\%$ of the training instances as the validation set.
\end{itemize}

Moreover, for the above three tasks, any pair of datasets for different tasks can be regarded as OOD to each other.
Besides, following \citet{hendrycks-etal-2020-pretrained}, we also select four additional datasets solely as the OOD data: concatenations of the premises and respective hypotheses from two \textbf{NLI} datasets \emph{RTE}~\cite{Dagan2005ThePR,BarHaim2006TheSP,Giampiccolo2007TheTP,Bentivogli2009TheSP} and \emph{MNLI}~\cite{Williams2018ABC}, the English source side of \textbf{Machine Translation (MT)} datasets English-German \emph{WMT16}~\cite{bojar2016findings} and \emph{Multi30K}~\cite{elliott2016multi30k}.
We take the test splits in those datasets as OOD instances in testing.
Particularly, for MNLI, we use both the matched and mismatched test sets.
For Multi30K, we use the union of the flickr 2016 English test set, mscoco 2017 English test set, and filckr 2018 English test set as the test set.
There are several reasons for not using them as ID data: (1) WMT16 and Multi30K are MT datasets and do not apply to a natural language classification problem.
Therefore, we cannot train a classifier on these two datasets.
(2) The instances in NLI datasets are labeled either as entailment/non-entailment for RTE or entailment/neural/contradiction for MNLI, which comprehensively covers all possible relationships of two sentences.
Therefore, it is hard to determine OOD instances for NLI datasets.
The statistics of the datasets are shown in~\Cref{tab:data_statistics}.

\begin{table}
    \centering
    \small
    \begin{tabular}{lcccc}
    \toprule
        Dataset & \# train& \# dev& \# test& \# class \\
    \midrule
        SST2& 67349& 872& 1821& 2 \\
        IMDB&22500& 2500& 25000& 2 \\
        TREC-10&4907&545&500&6 \\
        20NG& 15056& 1876& 1896& 20 \\
        \midrule
        MNLI&-&-&19643&- \\
        RTE&-&-&3000&- \\
        Multi30K&-&-&2532&- \\
        WMT16&-&-&2999&- \\
    \bottomrule
    \end{tabular}
    \caption{Statistics of the datasets.}
    \label{tab:data_statistics}
\end{table}

\begin{table*}[!t]
\centering
\renewcommand{\arraystretch}{1.0}
\small
    \begin{tabular}{llccccc}
    \toprule
    \multicolumn{2}{c}{\textbf{AUROC $\uparrow$ / FAR95 $\downarrow$}} &\textbf{Avg}& \textbf{SST2}& \textbf{IMDB}& \textbf{TREC-10}& \textbf{20NG} \\
    \midrule
    \multirow{4}{*}{\rotatebox[origin=c]{90}{w/o $\mathcal{L}_\text{cont}$}} & MSP& 94.1 / 35.0& 88.9 / 61.3 & 94.7 / 40.6& 98.1 / 7.6& 94.6 / 30.5 \\
    & Energy&94.0 / 34.7&87.7 / 63.2&93.9 / 49.5& 98.0 / 10.4& 96.5 / 15.8 \\
    & Maha& 98.5 / 7.3&96.9 / 18.3& 99.8 / 0.7& 99.0 / 2.7& 98.3 / 7.3 \\
    & Cosine& 98.2 / 9.7&96.2 / 23.6&99.4 / 2.1& 99.2 / 2.3& 97.8 / 10.7 \\
    \midrule
    \multirow{4}{*}{\rotatebox[origin=c]{90}{w/ $\mathcal{L}_\text{scl}$}} & $\mathcal{L}_\text{scl}$ + MSP&90.4 / 46.3& \colorit 89.7 / 59.9& 93.5 / 48.6& 90.2 / 36.4& 88.1 / 39.2 \\
    & $\mathcal{L}_\text{scl}$ + Energy& 90.5 / 43.5& 88.5 / 64.7& 92.8 / 50.4& 90.3 / 32.2& 90.2 / 26.8 \\
    & $\mathcal{L}_\text{scl}$ + Maha& 98.3 / 10.5&96.4 / 26.6&99.6 / 2.0& \colorit 99.2 / 1.9& 97.9 / 11.6\\
    & $\mathcal{L}_\text{scl}$ + Cosine& 97.7 / 13.0& 95.9 / 28.2&99.2 / 4.2& 99.0 / 2.4& 96.8 / 17.0\\
    \midrule
    \multirow{4}{*}{\rotatebox[origin=c]{90}{w/ $\mathcal{L}_\text{margin}$}} & $\mathcal{L}_\text{margin}$ + MSP&93.0 / 33.7&89.7 / 49.2&\colorit 93.9 / 46.3& 97.6 / 6.5& 90.9 / 32.6 \\
    & $\mathcal{L}_\text{margin}$ + Energy& 93.9 / 31.0& \colorit 89.6 / 48.8& \colorit 93.4 / 52.1& \colorit 98.4 / 4.6& 94.1 / 18.6 \\
    & $\mathcal{L}_\text{margin}$ + Maha& \colorit 99.5 / 1.7&\colorit 99.9 / 0.6&\colorit 100 / 0& \colorit 99.3 / 0.4& \colorit 98.9 / 6.0\\
    & $\mathcal{L}_\text{margin}$ + Cosine& \colorit 99.0 / 3.8& \colorit 99.6 / 1.7&\colorit 99.9 / 0.2& 99.0 / 1.5& 97.4 / 11.8 \\
    \bottomrule
    \end{tabular}
    \caption{OOD detection performance~(in \%) of RoBERTa$_{\textsc{LARGE}}$ trained on the four ID datasets. Due to space limits, for each of the four training ID dataset, we report the macro average of AUROC and FAR95 on all OOD datasets~(check Appendix for full results). 
    Results where the contrastive loss improves OOD detection on both evaluation metrics are highlighted in \colorbox{green!15}{green}.
    ``w/o $\mathcal{L}_\text{cont}$+MSP'' thereof is the method in \citet{hendrycks-etal-2020-pretrained}.
    }\label{tab::main_result}
\end{table*}

\subsection{Experimental Settings}\label{sec:exp_set}

\stitle{Evaluation Protocol.} We train the model on the training split of each of the four aforementioned ID datasets in turn. 
In the inference phase, the respective test split of that dataset is used as ID test data, while all the test splits of datasets from other tasks are treated as OOD test data.

We adopt two metrics that are commonly used for measuring OOD detection performance in machine learning research~\cite{hendrycks2016baseline, lee2018simple}:
(1) \textbf{AUROC} is the area under the receiver operating characteristic curve, which plots the true positive rate~(TPR) against the false positive rate~(FPR).
A \emph{higher} AUROC value indicates better OOD detection performance, and a random guessing detector corresponds to an AUROC of $50\%$.
(2) \textbf{FAR95} is the probability for a negative example~(OOD) to be mistakenly classified as positive~(ID) when the TPR is $95\%$, in which case a \emph{lower} value indicates better performance.
Both metrics are threshold-independent.

\stitle{Compared Methods.} We evaluate all configurations of contrastive losses and OOD scoring functions. 
Those include 12 settings composed of 3 alternative setups for contrastive losses ($\mathcal{L}_\text{scl}$,  $\mathcal{L}_\text{margin}$ or w/o a contrastive loss) and 4 alternatives of OOD scoring functions (MSP, the energy score, Maha, or cosine similarity).

\stitle{Model Configuration.}
We implement our framework upon Huggingface's Transformers~\cite{wolf-etal-2020-transformers} and build the text classifier based on RoBERTa$_{\textsc{LARGE}}$~\cite{liu2019roberta} in the main experiment.
All models are optimized with Adam~\cite{Kingma2015AdamAM} using a learning rate of $1\mathrm{e}{-5}$, with a linear learning rate decay towards 0.
We use a batch size of 32 and fine-tune the model for 10 epochs.
When training the model on each training split of a dataset, we use the respective validation split for both hyper-parameter tuning and 
The hyper-parameters are tuned according to the classification performance and the contrastive loss on the ID validation set.
We find that $\tau=0.3$ and $\lambda=2$ work well with $\mathcal{L}_\text{scl}$, while $\lambda=2$ work well with $\mathcal{L}_\text{margin}$, and we apply them to all datasets.

\begin{figure*}[!t]
    \centering
    \scalebox{0.81}{
    \includegraphics[width=\textwidth]{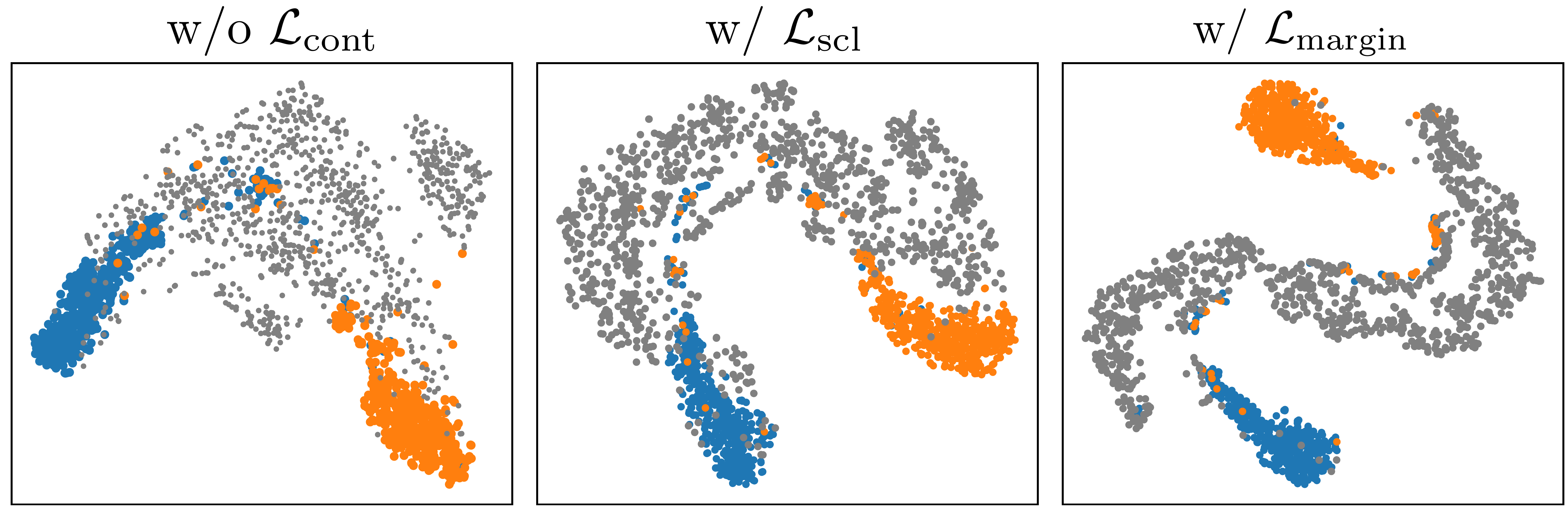}}
    \caption{Visualization of the representations for {\color{plot_orange}positive}, {\color{plot_blue}negative} instances in SST2 and {\color{plot_grey}OOD} ones.
    The discrepancy between ID and OOD representations is greater on representations obtained with $\mathcal{L}_\text{margin}$.}
    \label{fig:tsne}
\end{figure*}

\subsection{Main Results}\label{sec:results}

We hereby discuss the main results of the OOD detection performance. Note that the incorporation of our OOD techniques does not lead to noticeable interference of the main-task performance, for which an analysis is later given in \Cref{sec:analysis}.

The OOD detection results by different configurations of models are given in \Cref{tab::main_result}. 
For all results, we report the average of 5 runs using different random seeds.
Each model configuration is reported with separate sets of results when being trained on different datasets, on top of which the macro average performance is also reported.
For settings with $\mathcal{L}_\text{scl}$ and $\mathcal{L}_\text{margin}$, results better than the baselines~(w/o a contrastive loss) are marked as red.
We observe that:
(1) Among OOD detection functions, the Mahalanobis distance performs the best on average and drastically outperforms the MSP baseline used in~\citet{hendrycks-etal-2020-pretrained}. This is due to that the Mahalanobis distance can better capture the distributional difference.
(2) Considering models trained on different ID datasets, the model variants with $\mathcal{L}_\text{margin}$ have achieved near-perfect OOD detection performance on SST2, IMDB, and TREC-10.
While on the 20 Newsgroup dataset that contains articles from multiple genres, there is still room for improvement.
(3) Overall, The margin-based contrastive loss~($\mathcal{L}_\text{margin}$) significantly improves OOD detection performance.
Particularly, it performs the best with the Mahalanobis distance, reducing the average FAR95 of Maha by $77\%$ from $7.3\%$ to $1.7\%$.
(4) The supervised contrastive loss~($\mathcal{L}_\text{scl}$) does not effectively improve OOD detection in general.
In many cases, its performance is even worse than the baseline.

\begin{table}[!t]
\centering
\small
    \begin{tabular}{p{3.5cm}cc}
    \toprule
    \textbf{AUROC $\uparrow$ / FAR95 $\downarrow$}& \textbf{TREC-10}& \textbf{20NG} \\
    \midrule
    MSP& 73.7 / 56.5 & 76.4 / 80.7\\
    Maha& 75.5 / \textbf{56.1} & 77.2 / 74.1\\
    $\mathcal{L}_\text{margin}$ + MSP& 64.1 / 66.4& 74.6 / 82.0 \\
    $\mathcal{L}_\text{margin}$ + Maha& \textbf{76.6} / 61.3 & \textbf{78.5 / 72.7} \\
    \bottomrule
    \end{tabular}
    \caption{Novel class detection performance.}\label{tab::novel_class}
\end{table}

\subsection{Novel Class Detection}\label{secc:novel_class}
We further evaluate our framework in a more challenging setting of novel class detection.
Given a dataset containing multiple classes~($\ge 3$), We randomly reserve one class as OOD data while treating others as ID data.
We then train the model on the ID data and require it to identify OOD data in inference. 
In this case, the OOD data are sampled from the same task corpus as the ID data, and thus is much harder to be distinguished.
We report the average performance of 5 trials in~\Cref{tab::novel_class}.
The results are consistent with the main results in general.
The Mahalanobis distance consistently outperforms consistently outperforms MSP, and the $\mathcal{L}_\text{margin}$ achieves better performance except for the FAR95 metric on the TREC-10 dataset.
However, the performance gain is notably smaller than that in the main experiments.
Moreover, none of the compared methods achieve an AUROC score of over $80\%$.
This experiment shows that compared to detecting OOD instances from other tasks, detecting OOD instances from similar corpora is much more challenging and remains room for further investigation.

\subsection{Analysis}\label{sec:analysis}

\stitle{Visualization of Representations.}
To help understand the increased OOD detection performance of our method, we visualize the penultimate layer of the Transformer trained with different contrastive losses.
Specifically, we train the model on SST2 and visualize instances from the SST2 validation set and OOD datasets using t-SNE~\cite{van2008visualizing}, as shown in~\Cref{fig:tsne}.
We observe that 
the representations obtained with $\mathcal{L}_\text{margin}$ can distinctly separate ID and OOD instances, such that ID and OOD clusters see almost no overlap.

\stitle{Main Task Performance.}
As stated in~\Cref{ssec:task_definition}, the increased OOD detection performance should not interfere with the classification performance on the main task.
We evaluate the trained classifier on the four ID datasets.
The results are shown in~\Cref{tab::classification}.
We observe that the contrastive loss does not noticeably decrease the classification performance, nor does it increase the performance, which differs from the observations by~\citet{gunel2020supervised}.

\begin{table}[!t]
\centering
\small
    \begin{tabular}{p{2cm}cccc}
    \toprule
    \textbf{Accuracy}& \textbf{SST2}& \textbf{IMDB}& \textbf{TREC-10}& \textbf{20NG} \\
    \midrule
    w/o $\mathcal{L}_\text{cont}$&\textbf{96.4}& 95.3& \textbf{97.7}& 93.6 \\
    w/ $\mathcal{L}_\text{scl}$&96.3& 95.3& 97.4& 93.4\\
    w/ $\mathcal{L}_\text{margin}$& 96.3& 95.3& 97.5&\textbf{93.9}\\
    \bottomrule
    \end{tabular}
    \caption{Accuracy of the trained classifier.}\label{tab::classification}
\end{table}

\begin{table}[!t]
\centering
\scalebox{0.73}{
    \begin{tabular}{p{3.5cm}ccc}
    \toprule
    \textbf{AUROC $\uparrow$ / FAR95 $\downarrow$}& \textbf{L1}& \textbf{Cosine}& \textbf{L2} \\
    \midrule
    MSP& 93.6 / 31.1& \textbf{94.1 / 30.9}& 92.2 / 32.0 \\
    Energy& 93.8 / 27.2& \textbf{94.7 / 26.9}& 94.4 / 27.5\\
    Maha& 99.3 / 2.8& 99.2 / 3.0& \textbf{99.4 / 1.7}\\
    Cosine& 98.1 / 10.9& 98.8 / 5.3& \textbf{99.0 / 3.9}\\
    \bottomrule
    \end{tabular}}
    \caption{Average OOD detection performance of different distance metrics.}\label{tab::metrics}
\end{table}

\stitle{Distance Metrics.}
Besides L2 distance, we further evaluate the L1 distance and the cosine distance with the margin-based contrastive loss $\mathcal{L}_\text{margin}$.
Results are shown in~\Cref{tab::metrics}.
Due to space limitations, we only report the average OOD performance on the four ID datasets.
We observe that the three metrics achieve similar performance, and all outperform the baseline when using Maha as the scoring function.
Among them, L2 distance gets slightly better OOD detection performance.
Moreover, $\mathcal{L}_\text{margin}$ significantly outperforms $\mathcal{L}_\text{scl}$ when both use cosine as the distance metric.
It shows that their performance difference arises from the characteristics of the losses instead of the metric.

\begin{table}[!t]
\centering
\scalebox{0.78}{
    \begin{tabular}{p{3.4cm}cc}
    \toprule
    \textbf{AUROC $\uparrow$ / FAR95 $\downarrow$}& \textbf{Maha}& \textbf{Maha + $\mathcal{L}_\text{margin}$} \\
    \midrule
    BERT$_\textsc{BASE}$& 95.7 / 21.5& 98.4 / 8.1\\
    BERT$_\textsc{LARGE}$& 97.7 / 13.3& 99.1 / 3.9 \\
    RoBERTa$_\textsc{BASE}$& 98.4 / 9.3& 99.6 / 2.0 \\
    RoBERTa$_\textsc{LARGE}$& 98.5 / 7.3& 99.4 / 1.7\\
    \bottomrule
    \end{tabular}}
    \caption{Average OOD detection performance of other pretrained Transformers.}\label{tab::other_encoder}
\end{table}

\stitle{OOD Detection by Other Transformers.}
We also evaluate the OOD detection ability of other pretrained Transformers in~\Cref{tab::other_encoder} and report the average performance on the four ID datasets.
For BERT~\cite{devlin-etal-2019-bert}, we use $\lambda=0.2$.
We observe that:
(1) Larger models have better OOD detection ability.
For both BERT and RoBERTa, the large versions offer better results than the base versions.
(2) Pretraining on diverse data improves OOD detection.
RoBERTa, which uses more pretraining corpora, outperforms BERT models.
(3) The margin-based contrastive loss consistently improves OOD detection on all encoders.

\section{Conclusion}
This work presents an unsupervised OOD detection framework for pretrained Transformers requiring only ID data. We systematically investigate the combination of contrastive losses and scoring functions, the two key components in our framework. In particular, we propose a margin-based contrastive objective for learning compact representations, which, in combination with the Mahalanobis distance, achieves the best performance: near-perfect OOD detection on various tasks and datasets. We further propose novel class detection as the future challenge for OOD detection.

\section*{Ethical Consideration}
This work does not present any direct societal consequences. The proposed work seeks to develop a general contrastive learning framework that handles unsupervised OOD detection in natural language classification. We believe this study leads to intellectual merits that benefit with reliable application of NLU models. Since in real-world scenarios, a model may face heterogeneous inputs with significant semantic shifts from its training distributions. And it potentially has broad impacts since the tackled issues also widely exist in tasks of other areas. All experiments are conducted on open datasets.

\section*{Acknowledgment}

We appreciate the anonymous reviewers for their insightful comments and suggestions. 
This material is supported by the National Science Foundation of United States Grant IIS 2105329.

\bibliography{custom}
\bibliographystyle{acl_natbib}

\appendix

\clearpage
\onecolumn

\onecolumn

\onecolumn
\section{Full Results}
We show the full OOD detection performance of ID datasets on OOD datasets.
The results of w/o $\mathcal{L}_\text{cont}$ and w/ $\mathcal{L}_\text{margin}$ are shown in~\Cref{tab:cont_full} and~\Cref{tab:margin_full}, respectively.

\begin{table}
    \centering
    \scalebox{0.67}{
    \begin{tabular}{lcccccccccccccccc}
    \toprule
        \textbf{AUROC} &\multicolumn{4}{c}{\textbf{SST2}}&\multicolumn{4}{c}{\textbf{IMDB}} &\multicolumn{4}{c}{\textbf{TREC-10}}&\multicolumn{4}{c}{\textbf{20NG}} \\
        &MSP& Energy& Maha& Cosine& MSP& Energy& Maha& Cosine& MSP& Energy& Maha& Cosine& MSP& Energy& Maha& Cosine \\
        \midrule
        SST2& -& -& -& -&-&-&-&-&97.1& 94.8& 97.4& 97.9& 98.6& 99.6& 99.4& 99.7\\
        IMDB& -& -& -& -&-&-&-&-&98.9&98.8& 99.5& 99.5& 95.9& 97.8& 98.9& 98.6\\
        TREC-10& 91.8& 91.5& 97.8& 97.0& 94.9& 94.0&100& 99.5&-&-&-&-& 95.1& 97.6& 98.9& 98.7\\
        20NG& 93.6& 93.4& 94.9& 93.2& 96.0& 95.6& 99.8& 99.6& 98.2& 99.0& 99.5& 99.6&-&-&-&-\\
        MNLI& 84.6& 83.6& 95.1& 94.6& 93.1& 92.4& 99.5& 99.0&97.0&97.3& 98.9& 98.8& 94.1& 96.1& 98.1& 97.5\\
        RTE& 89.2& 87.4& 98.4& 98.1& 93.9& 93.3& 99.8& 99.5& 98.6&98.8& 99.5& 99.4& 90.3& 92.8& 96.5& 95.2\\
        WMT16& 84.0& 82.4& 96.4& 95.9& 93.4& 92.7& 99.7& 99.1& 97.9&98.2& 99.4& 99.3& 92.7& 95.0& 97.8& 96.8\\
        Multi30K& 90.2& 88.1& 98.8& 98.6& 96.4& 95.6& 99.9& 99.8& 99.1& 99.2& 99.7& 99.6& 95.7& 97.0& 98.7& 98.3\\
        \midrule
        Avg& 88.9& 87.7&96.9& 96.2& 94.7& 93.9& 99.8& 99.4&98.1& 98.0& 99.0& 99.2& 94.6& 96.5& 98.3& 97.8\\
        \bottomrule
    \end{tabular}}
\end{table}

\begin{table}
    \centering
    \scalebox{0.67}{
    \begin{tabular}{lcccccccccccccccc}
    \toprule
        \textbf{FAR95} &\multicolumn{4}{c}{\textbf{SST2}}&\multicolumn{4}{c}{\textbf{IMDB}} &\multicolumn{4}{c}{\textbf{TREC-10}}&\multicolumn{4}{c}{\textbf{20NG}} \\
        &MSP& Energy& Maha& Cosine& MSP& Energy& Maha& Cosine& MSP& Energy& Maha& Cosine& MSP& Energy& Maha& Cosine \\
        \midrule
        SST2& -& -& -& -&-&-&-&-&14.5& 28.5& 12.0& 6.4& 9.0& 2.3& 0.3& 0.9\\
        IMDB& -& -& -& -&-&-&-&-&2.9& 4.5& 0.2& 0.3& 25.3& 10.5& 4.4& 6.4\\
        TREC-10& 61.3& 63.1& 13.2& 19.4& 37.4& 57.7&0& 1.5&-&-&-&-& 35.0& 14.9& 1.3& 8.3\\
        20NG& 52.1& 52.5& 39.5& 55.9& 28.4& 32.1& 0.2& 0.6&7.2&6.6& 0.3& 1.0&-&-&-&-\\
        MNLI& 68.4& 68.7& 27.0& 31.4& 51.4& 55.4& 2.2& 4.8&13.6&15.5& 3.2& 4.2& 36.0& 20.2& 10.1& 13.9\\
        RTE& 59.7& 62.4& 8.0& 9.0& 49.9& 54.7& 0.7& 1.9&5.2&5.5& 0.8& 1.4& 49.0& 30.1& 17.1& 21.9\\
        WMT16& 69.4& 70.6& 17.2& 20.2& 50.9& 57.6& 1.2& 3.8& 8.5&10.2& 1.8& 2.2& 40.5& 23.1& 11.9& 16.4\\
        Multi30K& 57.3& 61.7& 5.5& 6.0& 25.7& 34.3& 0& 0.2& 1.3& 1.8& 0.3& 0.2& 18.9& 9.2& 5.9& 7.0\\
        \midrule
        Avg& 61.3& 63.2&18.3& 23.6& 40.6& 48.6& 0.7& 2.1& 7.6& 10.4& 2.7& 2.3& 30.5& 15.8& 7.3& 10.7\\
        \bottomrule
    \end{tabular}}
    \caption{AUROC and FAR95~(in \%) of RoBERTa$_{\textsc{LARGE}}$ model trained w/o $\mathcal{L}_\text{cont}$. Results are averaged over 5 runs with different seeds.}
    \label{tab:cont_full}
\end{table}

\begin{table}[!t]
    \centering
    \scalebox{0.67}{
    \begin{tabular}{lcccccccccccccccc}
    \toprule
        \textbf{AUROC} &\multicolumn{4}{c}{\textbf{SST2}}&\multicolumn{4}{c}{\textbf{IMDB}} &\multicolumn{4}{c}{\textbf{TREC-10}}&\multicolumn{4}{c}{\textbf{20NG}} \\
        &MSP& Energy& Maha& Cosine& MSP& Energy& Maha& Cosine& MSP& Energy& Maha& Cosine& MSP& Energy& Maha& Cosine \\
        \midrule
        SST2&-&-&-&-&-&-&-&-& 96.2& 96.6& 98.4& 97.8& 96.3& 98.1& 99.5& 99.0\\
        IMDB&-&-&-&-&-&-&-&-&99.3& 99.7& 99.6& 99.3& 94.5& 96.9& 99.0& 98.4\\
        TREC-10& 95.1& 94.9& 99.5& 99.0& 93.8& 93.3& 100& 100& -&-&-&-& 88.0& 92.4& 99.6& 96.5\\
        20NG& 95.2& 95.0& 100& 100& 95.4& 95.3& 100& 99.9& 99.2& 99.8& 99.8& 99.7&-&-&-&-\\
        MNLI& 82.8& 82.7& 99.8& 99.5& 92.4& 91.7& 100& 99.9& 96.6& 97.6& 99.2& 98.8& 91.0& 94.2& 98.4& 97.2\\
        RTE& 87.4& 87.5& 100& 99.9& 92.9& 92.1& 100& 99.9& 96.6& 98.1& 99.6& 99.2& 84.5& 88.7& 98.2& 95.6\\
        WMT16& 83.9& 84.0& 99.9& 99.4& 92.9& 92.2& 100& 99.9& 97.1& 98.0& 99.4& 99.1& 88.3& 92.5& 98.5& 96.7\\
        Multi30K& 93.5& 93.6& 100& 99.9& 95.9& 95.7& 100& 100& 97.9& 98.9& 99.5& 99.3& 93.7& 96.0& 99.1& 98.3\\
        \midrule
        Avg& 89.7& 89.6& 99.9& 99.6& 93.9& 93.4& 100& 99.9& 97.6& 98.4& 99.3& 99.0& 90.9& 94.1&98.9& 97.4\\
        \bottomrule
    \end{tabular}}
\end{table}

\begin{table}[!t]
    \centering
    \scalebox{0.67}{
    \begin{tabular}{lcccccccccccccccc}
    \toprule
        \textbf{FAR95} &\multicolumn{4}{c}{\textbf{SST2}}&\multicolumn{4}{c}{\textbf{IMDB}} &\multicolumn{4}{c}{\textbf{TREC-10}}&\multicolumn{4}{c}{\textbf{20NG}} \\
        &MSP& Energy& Maha& Cosine& MSP& Energy& Maha& Cosine& MSP& Energy& Maha& Cosine& MSP& Energy& Maha& Cosine \\
        \midrule
        SST2&-&-&-&-&-&-&-&-& 11.9& 10.4& 1.6& 6.9& 13.7& 5.3& 1.2& 2.6\\
        IMDB&-&-&-&-&-&-&-&-& 0.5& 0.2&0& 0& 23.6& 11.4& 4.7& 7.4\\
        TREC-10& 35.3& 35.0& 2.4& 4.3& 50.0& 54.0& 0& 0&-&-&-&-& 27.2& 13.8& 1.4& 4.4\\
        20NG& 36.4& 36.3& 0& 0& 37.8& 33.1& 0& 0& 0.6& 0.2& 0&0&-&-&-&-\\
        MNLI& 64.6& 64.3& 0.4& 2.6& 52.2& 83.8& 0.1& 0.9& 9.6& 6.7& 0.7& 1.9& 37.4& 24.7& 9.6& 16.7\\
        RTE& 58.3& 57.7& 0& 0.3& 52.9& 54.3& 0& 0.3& 9.8& 6.2& 0.1& 0.5& 52.9& 35.4& 11.1& 24.2\\
        WMT16& 64.3& 64.1& 0.5& 3.0& 53.7& 55.7& 0& 0.4& 7.9& 5.7& 0.5& 1.3& 45.3& 27.8& 7.5& 18.7\\
        Multi30K& 36.3& 35.4& 0& 0.3& 30.9& 31.9& 0& 0& 5.3& 2.6& 0& 0.2& 27.8& 12.0& 6.9& 8.7\\
        \midrule
        Avg& 49.2& 48.8& 0.6& 1.7& 46.3& 52.1& 0& 0.2& 6.5& 4.6& 0.4& 1.5& 32.6& 18.6& 6.0& 11.8\\
        \bottomrule
    \end{tabular}}
    \caption{AUROC and FAR95~(in \%) of RoBERTa$_{\textsc{LARGE}}$ model trained w/ $\mathcal{L}_\text{margin}$. Results are averaged over 5 runs with different seeds.}
    \label{tab:margin_full}
\end{table}
\end{document}